\documentclass{article}

     \usepackage[preprint]{neurips_2019}

\usepackage[utf8]{inputenc} 
\usepackage[T1]{fontenc}    
\usepackage{hyperref}       
\usepackage{url}            
\usepackage{booktabs}       
\usepackage{amsfonts}       
\usepackage{nicefrac}       
\usepackage{microtype}      

\usepackage{amssymb}
\usepackage{amsmath}
\usepackage{bm}
\usepackage{algorithm}
\usepackage{algpseudocode}
\usepackage{comment}
\usepackage{graphicx}

\title{Generating Explanations from Deep Reinforcement Learning Using Episodic Memory}

\author{%
  Sam Blakeman\\
  Sony AI \\
  Wiesenstrasse 5, Schlieren, 8952\\
  Switzerland \\
  \texttt{samrobertallan.blakeman@sony.com} \\
  \And
  Denis Mareschal \\
  Centre for Brain and Cognitive Development \\
  Department of Psychological Sciences\\
  Birkbeck, University of London \\
  Malet Street, WC1E 7HX UK \\
  \texttt{d.mareschal@bbk.ac.uk} \\
}

\begin{document}

\maketitle

\begin{abstract}
Deep Reinforcement Learning (RL) involves the use of Deep Neural Networks (DNNs) to make sequential decisions in order to maximize reward. For many tasks the resulting sequence of actions produced by a Deep RL policy can be long and difficult to understand for humans. A crucial component of human explanations is \textit{selectivity}, whereby only key decisions and causes are recounted. Imbuing Deep RL agents with such an ability would make their resulting policies easier to understand from a human perspective and generate a concise set of instructions to aid the learning of future agents. To this end we use a Deep RL agent with an episodic memory system to identify and recount key decisions during policy execution. We show that these decisions form a short, human readable explanation that can also be used to speed up the learning of naive Deep RL agents in an algorithm-independent manner.

\end{abstract}

\section{Introduction}

The combination of Deep Neural Networks (DNNs) \citep{lecun2015deep, schmidhuber2015deep} and Reinforcement Learning (RL), often referred to as Deep RL, has allowed artificial agents to solve human-like problems such as playing video games \citep{mnih2015human, silver2016mastering}. Despite this impressive achievement, the reliance of Deep RL on DNNs means that it is often criticised for being a 'black-box' approach that is unable to explain the behaviour that it has learnt. The purpose of the present research is to address this criticism by providing a method for generating concise task-level explanations using Deep RL.

The ability to explain how to solve a task allows humans to share learnt knowledge and speed up the collective learning process. A naive approach to generating an explanation would be to recall every decision made during the task. However, this is often undesirable because it leads to prohibitively long and complex explanations that cannot be easily understood by the recipient. It is therefore crucial that any explanation generating process is able to identify a small subset of key decisions that are fundamental for solving the task \citep{amir2019summarizing}. In the social sciences this is referred to as \textit{explanation selection} and refers to the fact that human explanations are biased to only a few important events or causes \citep{miller2019explanation}. Current approaches to generating explanations in Deep RL algorithms typically operate at the level of individual decisions, for example by computing saliency scores for all input features \citep{greydanus2018visualizing, mott2019towards, joo2019visualization}. They therefore do not produce selective task-level explanations and fundamentally lack \textit{selectivity} in their explanations \citep{alvarez2019weight}.

These approaches can be seen as focusing on 'why' decisions were made rather than 'what' the key decisions were to solve the task. However, to capture the complexity of human-level explanations both of these facets are important. For example, when asked to explain how to bake a cake we recount the key steps as an instructional list, leaving out minor details such as how to crack an egg or weigh the ingredients. This naive explanation does not involve causal reasoning but is still a valid explanation. Conversely, when asked to explain why a raising agent is added to the ingredients we reason about the chemical reactions that it causes in order to release carbon dioxide. Recent research has begun to tackle the former of these two types of explanation in Deep RL. In particular, \citep{guo2021edge} proposed a method for computing the importance of each time-step of an agent's trajectory for achieving reward. This is consistent with a selective task-level explanation because time-steps with high importance reflect key decisions made by the agent. However, the proposed approach requires the post-hoc training of a second explainable model from a set of episodes produced by the agent. The explanation generating process is therefore not an intrinsic part of the agent and an explanation cannot be generated from a single episode. In order to address these limitations we take inspiration from the brain to generate selective task-level explanations. 

One key brain structure involved in generating explanations is thought to be the hippocampus, which plays a central role in episodic and declarative memory \citep{eichenbaum1999hippocampus, eichenbaum2004hippocampus}. Indeed, it has been proposed that the primary role of episodic memory is to store memories of specific events so that they can be communicated to others \citep{dessalles2007storing, mahr2018we}. It therefore seems plausible that the addition of a hippocampal-like learning system may help Deep RL agents to generate task-level explanations. Several recent studies have explored the effect of adding a hippocampal learning system to Deep RL algorithms \citep{blundell2016model, pritzel2017neural, botvinick2019reinforcement, blakeman2020complementary, perrusquia2022complementary}. However, these studies have focused on the improved data efficiency that the hippocampal learning system provides \citep{gershman2017reinforcement} rather than its declarative properties. The question therefore arises; can the inclusion of a hippocampal learning system in Deep RL also provide a mechanism for generating selective task-level explanations?

To address this question we build on our previous work that outlined a framework for imbuing Deep RL algorithms with a hippocampal learning system \citep{blakeman2020complementary}. The approach, termed Complementary Temporal Difference Learning (CTDL), was inspired by the theory of Complementary Learning Systems (CLS) \citep{mcclelland1995there}. CLS theory states that the brain relies upon the complementary properties of the neocortex and the hippocampus to perform complex behaviour. In particular, the neocortex gradually learns overlapping representations from multiple experiences (i.e. semantic memory), while the hippocampus rapidly learns pattern-separated representations of individual experiences (i.e. episodic memory). As a result, these two learning systems allow for both a high-degree of generalization and rapid learning with minimal interference, which are both critical for efficient Reinforcement Learning (RL) behaviour \citep{botvinick2019reinforcement, gershman2017reinforcement}. CTDL provides support for this theory by demonstrating that the performance of classic Deep Reinforcement Learning (RL) algorithms can be improved by imbuing them with two learning systems that mimic the key properties of the neocortex and the hippocampus (also see \citet{blundell2016model, pritzel2017neural}). 

In CTDL, the neocortical learning system is represented by a Deep Neural Network (DNN) because of its ability to gradually learn overlapping representations from multiple experiences. Conversely, the hippocampal learning system is represented by a Self-Organizing Map (SOM) because individual memories are stored in a pattern-separated manner and can be updated quickly using a simple error-driven learning rule. Due to the limited memory capacities of the SOM the memories have to be stored and updated in a principled manner. To this end, CTDL uses the errors generated by the DNN to augment the unsupervised learning of the SOM. As a result, states of the environment that the DNN is poor at evaluating will be stored in the SOM along with an associated value estimate, which can then be updated in a tabular fashion. During action selection, the euclidean distance between the current state of the environment and the best matching memory in the SOM is used to calculate a weighted average between the value prediction of the SOM and the DNN. Intuitively, this results in a greater reliance on the predictions from the SOM if the state of the environment closely matches a memory stored in the SOM. CTDL can therefore use the DNN to generalize over large areas of the state space and the SOM to evaluate areas of the state space that are particularly hard to generalize over, for example where states have highly similar features but very different reward values.

This explicit communication between the neocortical and hippocampal learning system in CTDL is of interest for generating selective task-level explanations because it provides a mechanism for identifying key decisions based on the current task. We therefore propose that the content of the hippocampal learning system (i.e. the SOM) can be used to generate partial explanations of how to solve the current task. After learning the task, the memories that the agent uses from the SOM can be stored as a short ordered list of key states and actions. This list can then be interpreted as a partial explanation of how to solve the task and can be given to other agents to speed up their learning process. We demonstrate the efficacy of this approach in both the grid world and continuous mountain car domains. We visually explore the quality of the generated explanations and also perform a quantitative assessment by measuring the improvement in performance when a naive agent receives the explanation.

\section{Methods}

\subsection{Grid World and Continuous Mountain Car Tasks}

In the grid world task the environment is defined by a discrete two-dimensional grid. Each position in the grid is defined by two integer values, which correspond to the x and y position. The agent observes its discrete position in the grid and has to select one of four discrete actions; up, down, left and right. Each action moves the agent one step in the corresponding direction and results in a small negative reward of -0.05. An episode terminates either after 1000 steps or when the agent reaches the goal position for a positive reward of +1. A sub-sample of the grid positions are associated with a negative reward of -1 and therefore represent positions that should be avoided. As a result, the task of the agent is to reach the goal position in as few steps as possible while avoiding the negatively rewarded positions. An example of the task can be seen in Figure \ref{fig:maze_explanations}. If the agent tries to move outside of the grid then it remains where it is for that time-step. Agents were trained in an $\epsilon-greedy$ manner, with the probability of taking a random action linearly decreased from $1$ to $0.1$ over the first 200 episodes.

In addition to the grid world task, we also assess our approach on the continuous mountain car task. The continuous mountain car task is fundamentally different to the grid world task in that both the state and action space of the agent are continuous. The agent is represented by a car that starts in-between two hills and the goal of the agent is to reach the top of the right-hand hill. However the car's engine has insufficient power to drive directly up the hill and must therefore gather momentum in order to reach the top. The state of the agent is represented by two continuous values; one value represents the position of the car ($[-1.2, 0.6]$) and the other represents the velocity of the car ($[-0.07, 0.07]$). The action space of the agent is one dimensional and represents the force to be applied to the car ($[-1.0, 1.0]$) on that time-step. With respect to the reward function, the agent receives a positive reward of +100 for reaching the top of the hill as well as a negative reward that is equal to the squared sum of the actions chosen. The episode terminates when the car reaches the top of the hill or the episode length is greater than 1000. For all simulations we used the implementation of continuous mountain car provided by OpenAI gym \citep{brockman2016openai}. An example of the task can be seen in Figure \ref{fig:mountain_car_explanations}.

\subsection{Complementary Temporal Difference Learning (CTDL)}

\begin{figure}
	\centering
	\includegraphics[height=6.5cm, width=14cm]{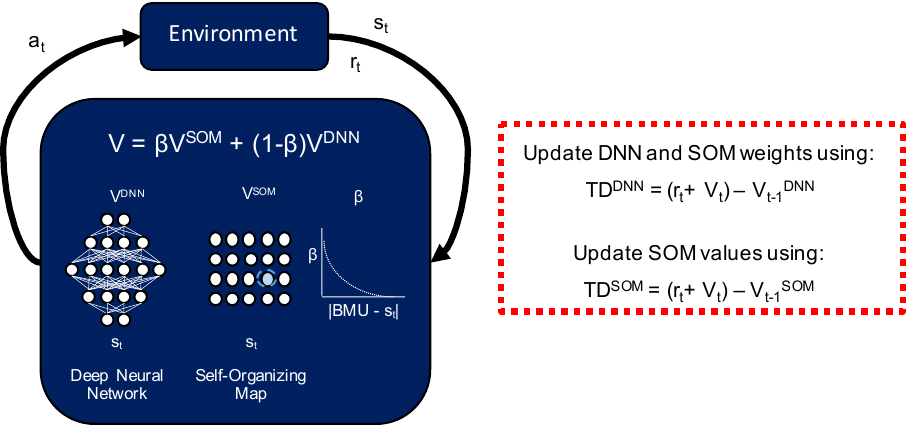}
	\caption{\textit{\textbf{Complementary Temporal Difference Learning (CTDL)}. CTDL consists of two main components; a Deep Neural Network (DNN) and a Self-Organizing Map (SOM). The DNN uses function approximation to predict the value ($V^{DNN}$) of the current state of the environment ($s_t$). In comparison, each unit in the SOM has a set of weights that corresponds to a state of the environment and stores an associated value estimate. The value prediction of the SOM ($V^{SOM}$) is the value associated with the unit whose weights are closest to $s_t$ i.e. the Best Matching Unit (BMU). During action selection, a weighting value ($\beta$) is calculated based on the Euclidean distance between $s_t$ and the weights of the BMU. $\beta$ is then used to calculate a weighted average ($V$) of the predictions made by the SOM and the DNN. During training (red box) the TD errors for both the DNN ($TD^{DNN}$) and the SOM ($TD^{SOM}$) are calculated. $TD^{DNN}$ is used to update the weights of both the DNN and the SOM, while $TD^{SOM}$ is used to update the values of the SOM in a tabular fashion. Crucially this architecture means that the SOM is biased towards storing states that the DNN is poor at predicting.}}
	\label{fig:ctdl}
\end{figure}

In this section we briefly describe the key components of Complementary Temporal Difference Learning (CTDL) (Figure \ref{fig:ctdl}), however for an in-depth description of CTDL please see \citet{blakeman2020complementary}. The fundamental components of CTDL are a Deep Neural Network (DNN), which represents a neocortical learning system, and a Self-Organizing Map (SOM), which represents a hippocampal learning system. The DNN can be any standard Deep RL algorithm involving a value function, such as a Deep Q-Network \citep{mnih2015human} or Advantage Actor-Critic \citep{mnih2016asynchronous}. However the final value prediction is not solely based on the prediction of the DNN but is instead a weighted average of the predictions from the DNN and the SOM.

Each unit in the SOM has a set of weights that corresponds to a state of the environment. These weights are updated using the standard learning rules of a SOM. However, the width of the neighbourhood function and the size of the learning rate is scaled by the Temporal Difference (TD) error produced by the value prediction of the DNN. In this way the SOM is updated to represent states of the environment that the DNN is poor at evaluating. In addition each unit of the SOM has a single value estimate, which corresponds to the predicted value of the state represented by its weights. This value estimate is updated in a simple tabular fashion with a high learning rate.

The final value prediction of CTDL is a weighted average of the value predictions from the DNN and the closest matching unit in the SOM. This weighting value ($\beta$) is calculated by passing the negative Eucliden distance between the current state and the closest matching state in the SOM through an exponential function. As a result, the closer the current state is to the closest matching unit in the SOM the more CTDL will rely on the prediction from the SOM. The subsequent weighted average of the predictions from the SOM and the DNN is used to train both components and also to select actions if a policy is not explicitly learnt (e.g. in the case of DQN).

\subsection{Generation of Explanations from CTDL}

After learning the task, the memories stored in the hippocampal learning system (i.e. the SOM) of CTDL contain a record of the states of the environment, and their associated values, that were particularly difficult for the DNN to evaluate. They therefore represent key decision points of the task where rapid learning and reduced interference between states is critical. With this in mind, we propose to generate partial explanations at the task-level by selecting a subset of these memories and presenting them as a temporal sequence.

\begin{figure}
	\centering
	\includegraphics[height=6cm, width=14.5cm]{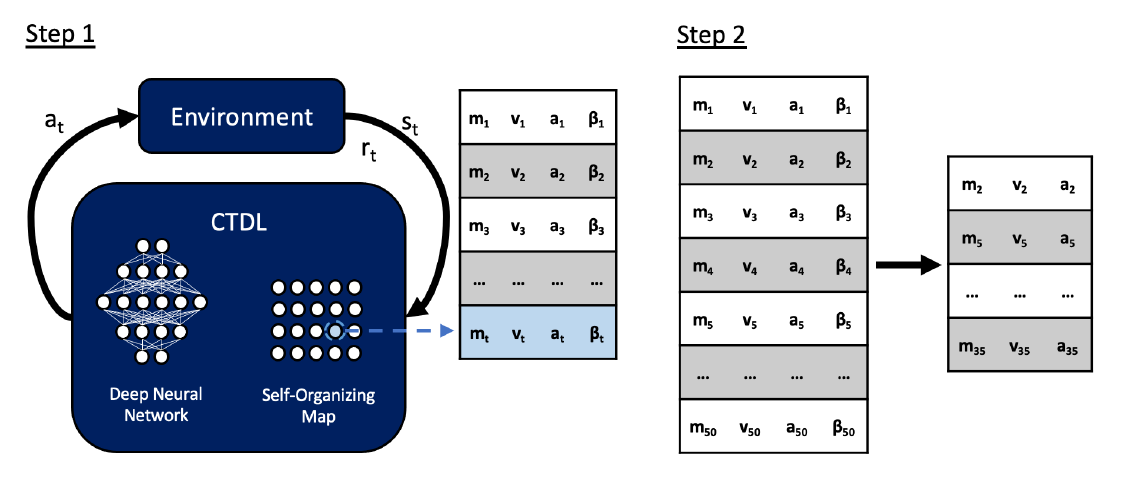}
	\caption{\textit{\textbf{Process of generating explanations from Complementary Temporal Difference Learning (CTDL). Step 1:} After training an agent via CTDL, a test trial is performed. During the test trial, an ordered list is kept of all the memories used from the Self-Organizing Map (SOM). In addition to the memory (m), the value associated with that memory (v), the degree to which the value was used ($\beta$) and the action taken (a) are also recorded. \textbf{Step 2:} After the test trial has been completed the list is pruned to provide a partial explanation of how to solve the task. For each unique memory in the list, only the row with the highest value of $\beta$ is kept. This ensures that each memory only has a single associated value and action. In addition, all rows where $\beta<0.5$ are removed as they formed the minority of the value prediction and so were not heavily relied upon by the agent.}}
	\label{fig:generating_explanations}
\end{figure}

In order to make this selection, we ask the agent to perform a test trial at the end of learning. During this test trial no further learning occurs and we keep a list of every memory that was used from the SOM along with its associated tabular value. This equates to keeping a record of the memory in the SOM that is closest to the current state based on Euclidean distance on each time-step. We also keep a record of the action that was taken and the calculated weighting value ($\beta$) for each memory. As a reminder, $\beta$ is calculated at each time-step using the Euclidean distance between the current state and the closest matching memory in the SOM. It is then used during the state evaluation process to calculate a weighted average between the prediction of the SOM and the DNN. 

Recording $\beta$ is important for reducing the length of the list post-hoc so that the explanation is more concise and understandable. Firstly, during the test trial the same memory in the SOM may be used multiple times by the agent but the associated action may be different each time. We therefore use $\beta$ to keep the memory-action pairing that had the biggest contribution to the state evaluation process. This ensures that a memory only has one entry in the list and therefore one associated action. Secondly, we only keep memories in the list where $\beta>0.5$ because this means that the memory formed the majority of the value prediction. This makes sure that we do not include memories in the explanation that contributed very little to the state evaluation and therefore the behaviour of the agent.

This process can also be performed online during the test trial. Memories, values, actions and $\beta$ values can be appended to the list whenever $\beta>0.5$. If a memory already exists in the list then it is removed whenever a higher $\beta$ value is encountered, and a new entry is appended to the list. We therefore do not need to store values for every transition on the test trial in order to generate an explanation.

Whether online or offline, the overall result is a list where each entry has a memory has an associated value and action. Each entry in the list has a unique memory and the list is ordered to reflect the temporal order in which the agent used the memories. This list can be easily visualised to produce a visual explanation of how to solve the task or it can be given to other agents to speed up their learning process. Figure \ref{fig:generating_explanations} shows a graphical depiction of the whole explanation-generating process.

\subsection{Provision of Explanations}

After generating an explanation from CTDL, the list of memories, values and actions can be provided to other agents to improve the efficiency of their learning. In order to utilise the list the receiving agent simply needs to calculate the weighting value $\beta$ between the current state of the environment and the memories in the list on each time-step. If the weighting is greater than a predefined threshold (e.g. $0.5$) for a memory in the list then the agent's current action and value estimate can be set to that memories action and value. If multiple memories have a weighting greater than the threshold then the one with the highest value is used. The benefits of this simple mechanism are two fold; (1) the policy is guided towards critical actions early on and (2) RL algorithms that use a value function can use the associated values to bootstrap value estimates during learning.

While this mechanism of providing explanations can be used for any RL algorithm, we can enhance it further if the explanation is being provided to CTDL. In this case, the list of memories, values and actions can be used to randomly initialise the entries of the SOM. These entries are fixed throughout the course of learning so that they are not overwritten. CTDL can then use these entries to further guide action selection and bootstrap learning. This naturally gives us a way to transfer the explanation into episodic memory before learning begins. Importantly, it also highlights that different algorithms can use the explanations in different ways depending on their underlying architecture.

\section{Results}

\subsection{Generating Explanations}

\begin{figure}
	\centering
	\includegraphics[height=6cm, width=14cm]{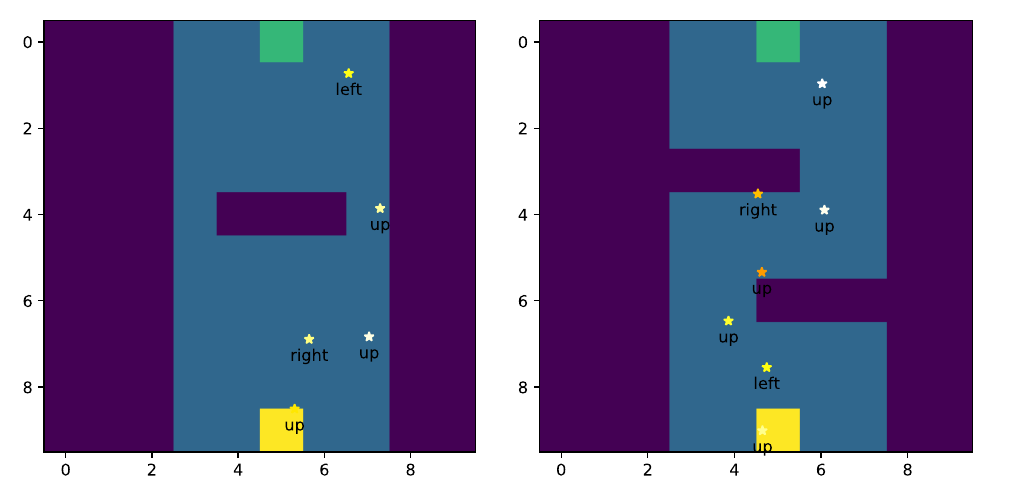}
	\caption{\textit{\textbf{Example explanations generated from Complementary Temporal Difference Learning (CTDL) on two different grid worlds.} The agent starts on the yellow square and has to move to the green square, which is associated with a reward of +1. The dark blue squares are associated with a reward of -1 and every action causes a reward of -0.05. The explanation is represented by stars, which correspond to memories extracted from the Self-Organizing Map (SOM). To generate an explanation, an agent performs a test trial after training. During the test trial, if a memory in the SOM is used with a weighting greater than 0.5 then it is recorded along with the action that was taken. If at the end of the test trial the same memory was used multiple times then the instance with the highest weighting is kept. This results in a short list of memories and actions, which provides a partial explanation of how to solve the given problem.}}
	\label{fig:maze_explanations}
\end{figure}

For the grid world experiments, we trained 12 agents for 1000 episodes on two different grid worlds and generated an explanation after every 200 episodes. Figure \ref{fig:maze_explanations} shows the explanations extracted from the best performing agent after 1000 episodes of training. The best agent was the one that achieved the most reward on a test trial after training. If a tie existed then the agent with the highest training reward was chosen. Since an explanation is simply a list of state-action pairings it can easily be inspected and qualitatively assessed. From visual inspection, the explanations include the essential decisions needed to solve each grid world. Crucially, the explanations do not include every action taken by the agent, which demonstrates that the explanation mechanism is able to select only the most important state-action pairings. Figures \ref{fig:grid_world_all_1} and \ref{fig:grid_world_all_2} in the Supplementary Materials show the performance and explanations of all 12 agents.

\begin{figure}
	\centering
	\includegraphics[height=19cm, width=12cm]{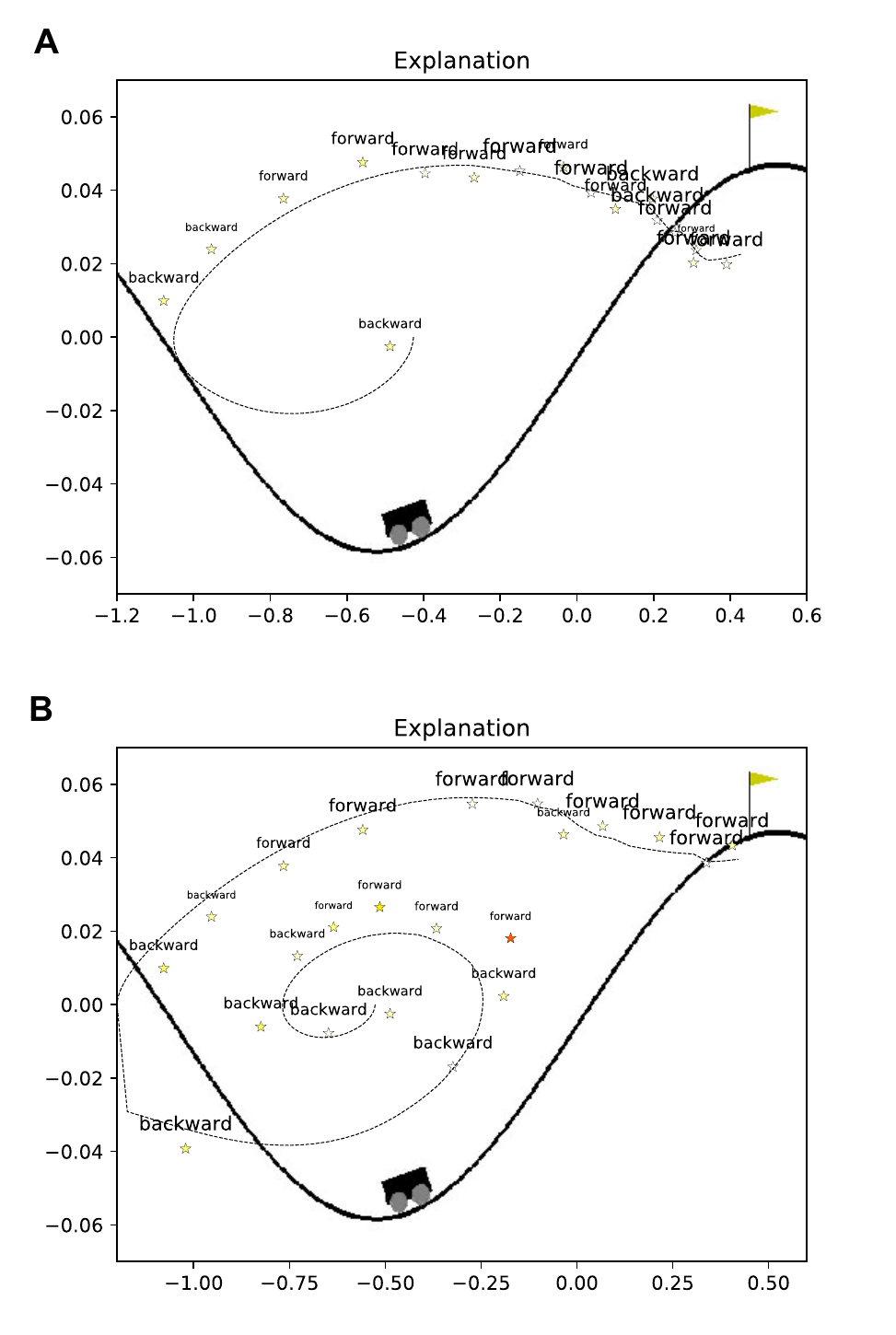}
	\caption{\textit{\textbf{Example explanations generated from Complementary Temporal Difference Learning (CTDL) on the continuous mountain car problem.} The agent has to gather momentum in order to escape from the valley and reach the flag for a reward of XXX. The x-axis represents the position of the car and the y-axis represents the velocity. The dashed line indicates the trajectory of the car for a single test trial after learning. Both trajectories are generated by the same agent but are different due to stochastic action selection and starting positions of the car. Explanations were generated in the same way as in the grid world experiments. Each star represents an individual memory, and its associated action is displayed above. The size of the text indicates the magnitude of the action taken.}}
	\label{fig:mountain_car_explanations}
\end{figure}

For the continuous mountain car task, 50 agents were trained for 1000 episodes and explanations were generated at the very end of training. Figure \ref{fig:mountain_car_explanations} shows the explanations extracted from the best performing agent. As with the grid worlds, the explanations do not involve every decision made by the agent but instead represent key decisions for solving the task. Interestingly, both of the explanations in Figure \ref{fig:mountain_car_explanations} were generated by the same agent. The reasons for these differences is due to the stochastic action selection and the randomized starting point of the car. In Figure \ref{fig:mountain_car_explanations}A the agent starts nearer to the goal and so a single backward action can gather enough momentum to then accelerate forward and reach the goal. In comparison, in Figure \ref{fig:mountain_car_explanations}B the agent starts further from the goal and so more traverses of the valley are needed to gain the momentum to reach the goal. In both cases an explanation can be extracted to represent these two different strategies. 

The length of the generated explanations can be reduced by increasing the $\beta$-threshold that is used to prune the list of episodic memories (Figure \ref{fig:generating_explanations}). Increasing the $\beta$-threshold ensures that only episodic memories that were heavily relied upon during policy execution form part of the explanation. Figure \ref{fig:explanation_threshold} in the Supplementary Materials demonstrates how the length of the explanation is reduced when the $\beta$-threshold is increased for the continuous mountain car task.



\subsection{Providing Explanations}

To test quantitatively how useful the generated explanations were, we provided them to new agents to see whether they improved learning. The original agents formed a baseline measure of performance because they did not receive an explanation during training and were only used to generate the explanations. We refer to these original agents as the no explanation group. In comparison, the new set of agents received explanations from the best performing original agent at the start of learning and so we refer to them as the explanation group.

\begin{figure}
	\centering
	\includegraphics[height=9cm, width=14cm]{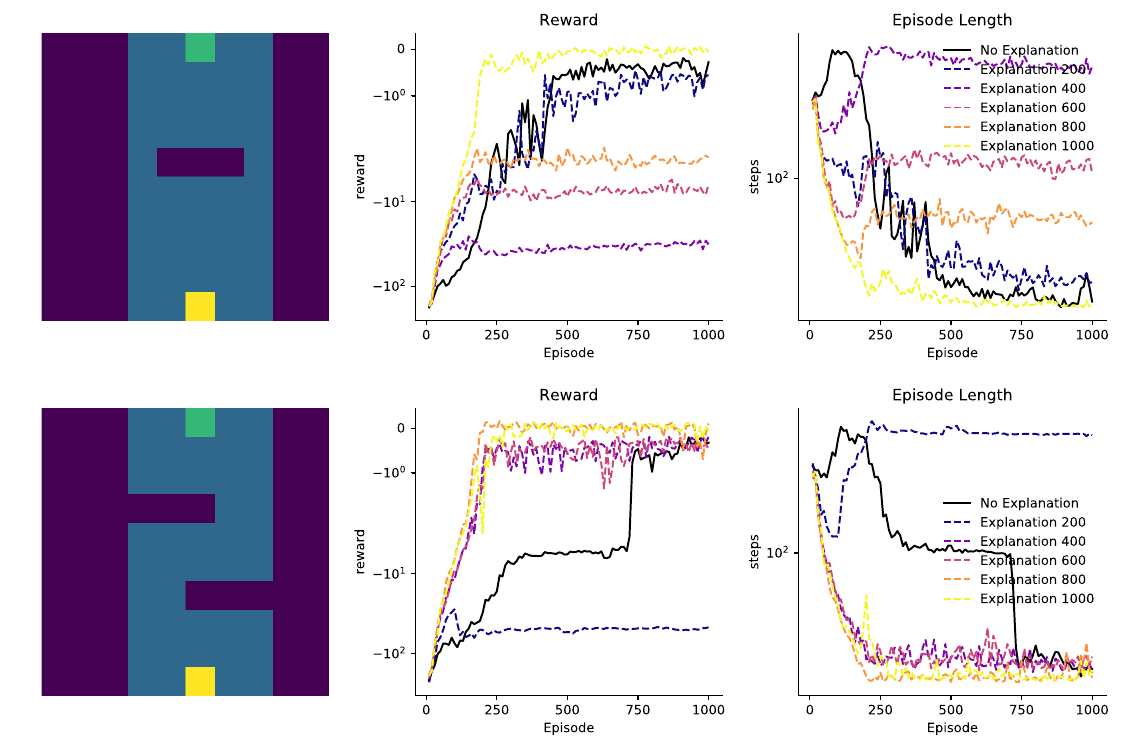}
	\caption{\textit{\textbf{The performance on two different grid worlds of the original agents (No Explanation) and the agents that received an explanation generated from the best original agent at different points of learning (Explanation 200, 400, 600, 800 and 1000).} The best performing original agent generated an explanation every 200 episodes. This created five different explanations and each one was provided to a new group of agents at the start of learning. To the right of each grid world is the reward achieved and the length of the episode over the course of learning for all groups. Each line is an average over 12 agents and each point is averaged over 5 episodes to improve visibility.}}
	\label{fig:grid_world_time_comparisons}
\end{figure}

For the grid world simulations, the explanations from the best original agent were generated every 200 episodes during training. We therefore trained 5 groups of 12 new agents, with each group receiving an explanation from a different point in the original agents training. This allowed us to explore whether explanations generated later on in training were more useful than those generated earlier on in training. Figure \ref{fig:grid_world_time_comparisons} compares the performance of the 5 new groups of agents as well as the original best performing agent. Most strikingly, agents who received an explanation generated from the best original agent at the end of training (after 1000 episodes) were able to solve both grid worlds in only $\sim$250 episodes. This represents a significant improvement in learning efficiency and demonstrates the utility of the generated explanations. When looking at explanations generated from different points in the original agent's training, we generally saw that explanations generated late on in training were more useful. Interestingly, agents that received an explanation from earlier points in training of the original agent did also demonstrate rapid learning but their asymptotic performance was typically reduced. This suggests that explanations generated from agents that have not converged to the optimal solution can in fact be misleading and detrimental to performance. One way to interpret this is that experts provide the best explanations and explanations from novices can limit one's ability to discover a more optimal solution.

\begin{figure}
	\centering
	\includegraphics[height=9cm, width=14cm]{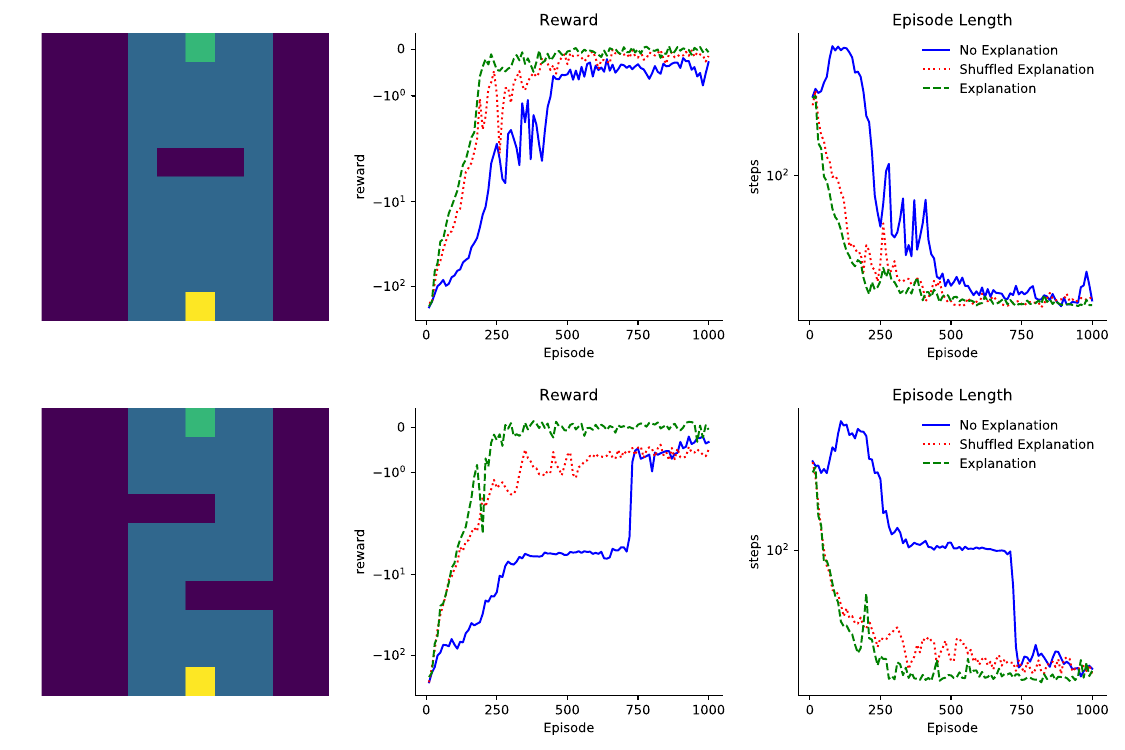}
	\caption{\textit{\textbf{The performance on two different grid worlds of the original agents (No Explanation), the agents that received an explanation generated from the best original agent at the end of learning (Explanation), and the agents that received a random sample of the memories generated by the best original agent at the end of learning (Shuffled Explanation).} To the right of each grid world is the reward achieved and the length of the episode over the course of learning for all groups. Each line is an average over 12 agents and each point is averaged over 5 episodes to improve visibility.}}
	\label{fig:grid_world_shuffled_comparisons}
\end{figure}

Having demonstrated the utility of explanations generated at the end of learning, we next investigated the importance of the pruning process for generating the explanations (Step 2 in Figure \ref{fig:generating_explanations}). To this end, we simulated a third group that we refer to as the shuffled explanation group. This group received an explanation that was generated by selecting a random subset of the memories used by the best agent on the test trial after training. This is equivalent to taking a random sample in Step 2 of Figure \ref{fig:generating_explanations} rather than pruning based on how much the original agent relied on each memory. This group therefore allowed us to assess how effective our pruning step was compared to taking a random sample. For a fair comparison, the size of the random sample is chosen to match the number of memories left after pruning. Figure \ref{fig:grid_world_shuffled_comparisons} shows the performance of the no explanation, explanation (generated after 1000 episodes) and shuffled explanation groups. The shuffled explanation group performed slightly worse than the explanation group in both gird worlds in terms of speed of learning and asymptotic performance. However, the shuffled explanation group still learnt significantly faster than the no explanation group. This suggests that our method of pruning memories to construct an explanation does confer some benefit over randomly sampling the memories used during the test trial. However, we shall see in the results of the continuous mountain car task that for more complex problems the utility of the pruning process appears to increase.

The aforementioned results demonstrate the our approach to generating partial task-level explanations from CTDL can confer substantial learning benefits in grid world environments. We therefore wanted to explore whether this benefit was also present in more complex environments. To this end, we evaluated our approach on the continuous mountain car task, which involves both continuous states and actions. Importantly, explanations were generated from the best original agent at the end of learning (after 1000 episodes) in exactly the same way as the grid world simulations.

\begin{figure}
	\centering
	\includegraphics[height=7.25cm, width=14.5cm]{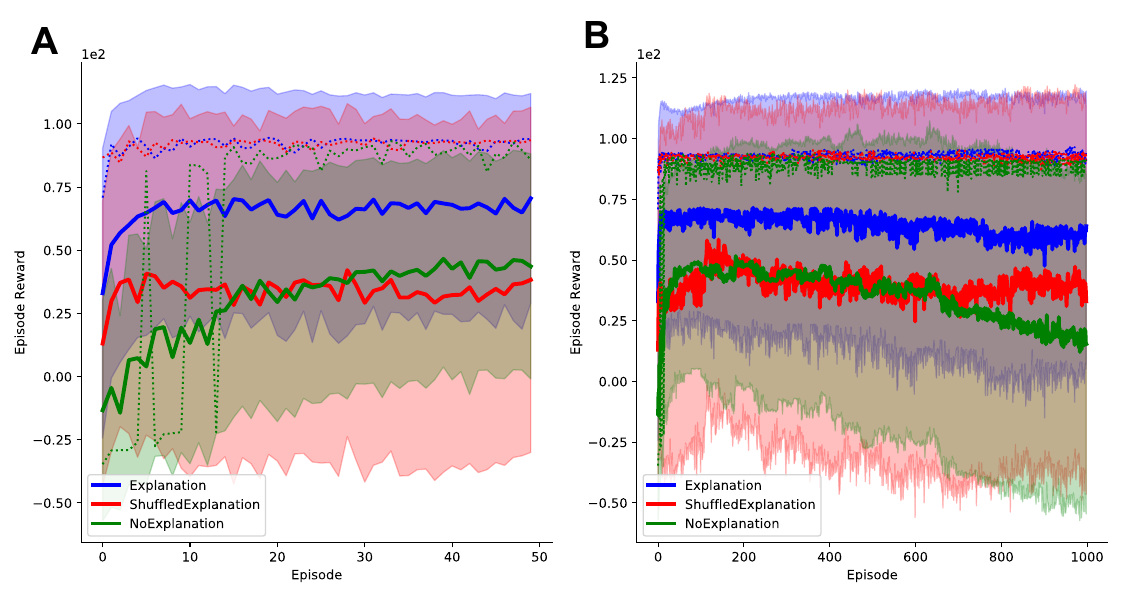}
	\caption{\textit{\textbf{The performance on the continuous mountain car task of the original agents (No Explanation), the agents that received an explanation generated from the best original agent at the end of learning (Explanation), and the agents that received a random sample of the memories generated by the best original agent at the end of learning (Shuffled Explanation).} 50 agents were trained on the continuous mountain car task for 1000 episodes. The agent with the highest total reward on the final episode was chosen to provide explanations. Explanations were generated by running the chosen agent on 20 test episodes after training. The explanations were then used to train 50 new agents with each new agent picking one at random. Solid lines indicate the average performance over 50 agents. Dashed lines indicate the best performing agent for each group. \textbf{(A)} Performance on the first 50 episodes of training. \textbf{(B)} Performance on all 1000 episodes of training.}}
	\label{fig:mountain_car_comparisons}
\end{figure}

Figure \ref{fig:mountain_car_comparisons} compares the performance of the no explanation, explanation and shuffled explanation groups on the continuous mountain car task. When comparing the no explanation and explanation groups, agents that received an explanation achieved higher levels of reward on average than those that did not. Importantly, the provision of an explanation did not appear to lead to the discovery of a better overall policy since the best performing agents in both cases reached a similar level of performance (see the dashed lines in Figure \ref{fig:mountain_car_comparisons}). This is to be expected given that the provided explanations describe the strategies learnt by the agents without an explanation and so in both cases the policies should be qualitatively similar. The provision of an explanation therefore appears to increase the probability of an agent finding a previously learnt policy rather than discovering a new optimal policy. As the explanations are generated from the best original agent, the agents receiving the explanation benefit from the increased probability of finding this best policy and so the average performance of the overall population increases.

Upon closer inspection of the first 50 episodes of training (Figure \ref{fig:mountain_car_comparisons}A), agents that received an explanation not only found a better policy on average but also converged to that policy faster. This is demonstrated by the fact that the episode reward for agents that received an explanation increased much more rapidly in the first few episodes of training compared to agents that did not receive an explanation. This was even the case when comparing to the best performing agent that did not receive an explanation. In addition, from looking over all 1000 episodes of training (Figure \ref{fig:mountain_car_comparisons}B), agents that received an explanation also appeared to learn a more robust policy. The average performance of the agents without an explanation appears to decrease after an initial increase in learning. In comparison, the average performance of the agents with an explanation appears to stay relatively stable after the initial increase. This may be due to the fact that the key decision points that are provided are not updated and so they provide a rigid scaffold for the policy that cannot change.

When looking at the shuffled explanation group in Figure \ref{fig:mountain_car_comparisons}, the average performance was substantially worse than the explanation group. This difference is strikingly larger than in the grid world experiments where the difference between the explanation and shuffled explanation groups was only marginal. We hypothesize that this is due to the larger continuous state space of the continuous mountain car task. As a result, the randomly selected memories of the SOM cannot provide an extensive coverage of the problem and so a targeted pruning process based on how much the agent uses the memories becomes increasingly important.

In summary, the aforementioned results demonstrate quantitatively that the explanations generated from CTDL can be understood and utilized by a naive CTDL agent. More specifically, receiving an explanation from CTDL confers several clear benefits; an increased probability of converging to the policy used to generate the explanation, faster convergence to this policy and increased robustness of the final policy. Crucially, the explanation should be generated by an agent that has learnt the task and converged to a desirable solution. This makes intuitive sense as explanations from individuals who have not yet grasped a task are likely to be misleading and potentially detrimental to one's own performance.

These results demonstrate the utility of the explanations when they are provided to a naive CTDL agent. However, because the explanations are a simple list of states, values and actions, they can be used by any RL algorithm. In order to demonstrate this, Figure \ref{fig:mountain_car_A2C_comparisons} in the Supplementary Materials shows the performance of Advantage Actor-Critic (A2C) \citep{mnih2016asynchronous} both with and without an explanation from CTDL. As with the naive CTDL agents, the average reward was higher and the policy appeared to be more robust when the naive A2C agent received an explanation at the start of training. This illustrates that our approach is algorithm-independent and therefore highly flexible.

\section{Discussion}

Explanations of how to solve a task often involve a summary of the key decisions required to complete it, an ability referred to as \textit{selectivity} \citep{miller2019explanation, alvarez2019weight}. Classic Deep Reinforcement Learning (RL) approaches lack this ability because all actions are reported when executing a policy. Recently, Complementary Temporal Difference Learning (CTDL) has been proposed which uses a Deep Neural Network (DNN) and a Self-Organizing Map (SOM) to solve the RL problem. Importantly, CTDL uses the errors produced by the DNN to update the contents  of the SOM. In effect this results in the SOM storing episodic memories of states and actions that led to the largest errors during learning. We therefore use the contents of the SOM to generate task-level explanations as they provide an intuitive summary of most the important state-action pairs for solving the task at hand.

From a qualitative perspective, the explanations generated from CTDL appeared to capture the critical structure of the current task. In the grid world experiments, the sequence of states and actions can be followed in order to trace a route to the goal location but they do not exhaustively cover the whole trajectory. Similarly, in the continuous mountain car task, the basic strategy of gaining momentum can be easily seen from the generated explanation without the need to report every action taken. The explanations therefore gave us a condensed view of the strategy learnt by the agent in an understandable and human-readable format.

In order to obtain a quantitative assessment of the explanations generated from CTDL, we also provided them to naive agents at the start of learning to see whether they improved performance. In the case of both the grid worlds and the continuous mountain car task, we saw better average performance, faster learning and increased robustness when an explanation was provided to the agent. However, this was only true when the explanation was generated at the end of learning i.e. by an expert. Explanations from early on in learning often appeared to be misleading and limit the final performance of the agent that received it. This appears to be in line with human explanations, whereby explanations based on false beliefs can lead to further inaccuracies \citep{lombrozo2006structure}.

While the SOM of CTDL contains memories responsible for large prediction errors during learning, not all of them may be useful for the final policy/strategy found by the agent. We therefore used a pruning process to ensure that the extracted explanation only used memories from the SOM that were used by the agent at test time. We found that this pruning process was particularly useful for the continuous mountain car task, where it provided a substantial benefit over just randomly selecting memories from the SOM to form an explanation. We believe this is due to the continuous state space of the continuous mountain car task, which is substantially larger than the small discrete state space of the grid world experiments. As a result, the memories stored in the SOM can represent very different regions of the state space and not all of them may be visited after convergence to the final policy. 

There are several reasons why our method of providing task-level explanations to naive agents resulted in substantial learning benefits. Firstly, the explanations were generated from the best original agent and so the new agents instantly benefit from the exploration that was performed at the population level. If an explanation had been provided by a worst original agent then the new agents will be biased towards discovering the same poorly performing policy. Support for this comes from the fact that agents performed poorly when they received an explanation that was generated early on in the original agent's training.

Secondly, by receiving an explanation the learning problem faced by the agent is made substantially easier because fewer state-action mappings have to be learnt. More specifically, the agent only has to learn the actions in-between the key decision points provided by the explanation. Learning to fill in these gaps is still important because agents that received an explanation still showed an increase in performance over the course of learning. However, this is a substantially easier learning problem compared to learning an entirely new policy. 

Another benefit of receiving an explanation containing key decision points is that that the agent experiences positive rewards sooner. This results in a stronger training signal early on in learning, which is particularly beneficial when rewards are sparse. This is demonstrated by the fact that agents that received an explanation in the continuous mountain car task achieved substantially more reward on the very first episode compared to those that did not receive an explanation.

Finally, the generated explanations not only contain state-action pairings but also their associated values. Crucially, these values are relative to the best policy and so they can be quickly bootstrapped to speed up the learning of other values. Indeed, this may be why we saw an improvement in performance even when memories were randomly sampled from the SOM to form an explanation. The actions themselves may not be used for the final policy but their associated values can still be used to improve the value estimates of states and actions that do form the final policy.

One way to summarise these benefits is that the provision of an explanation helps to simplify the exploration problem faced by the naive agent. The exploration problem is simplified because the naive agent can use the key decisions found by another agent, as well as their associated values, to scaffold their learning. This is particularly important for tasks such as the continuous mountain car task, where positive rewards must be experienced early on in training or the agent will learn to minimise the sum of its actions and not move.

We have shown that this simple mechanism for generating task-level explanations can be used to provide explanations to any RL algorithm. The explanations generated by CTDL consist of a simple list of states, actions and values, which form the basis of most Deep RL algorithms. For a Deep RL algorithm to use the list during learning, all it requires is a kernel function that decides when to use an entry from the list. In the present work we used the Euclidean distance between the current state and the states in the list to make this decision based on a pre-defined threshold. However, this kernel function could take other potentially more useful forms. One interesting alternative is the Successor Representation (SR) \citep{dayan1993improving, momennejad2017successor}, which calculates future state occupancies based on the current state. Based on the SR, if a state in the list is highly likely to be visited in the near future then the agent should use the associated value and action form the explanation. Future work should therefore explore the utility of different kernel functions for deciding when to use the contents of the explanation.

One important limitation of our proposed approach is that the list of states and actions generated are in the same space as experienced by the agent. In the case of the grid world and continuous mountain car tasks these states and actions could be visualized using 2-dimensional plots. If the state space of the agent was based on images then this would still not be a problem because the explanation would generate a small subset of images that could be labelled with their corresponding actions. However, some tasks may involve high-dimensional state and action spaces that cannot be easily visualized for humans. This would limit the impact of the generated explanations and another system would be required to transform the list of states and actions into a lower dimensional and human readable representation. One possibility is to combine our approach with those that report individual feature saliency. This would reduce the dimensionality of the states in the list because only the most important features of each key decision could be visualized.

\section{Concluding Remarks}

We have proposed a novel method for generating selective task-level explanations from Deep RL agents. The explanations summarize the key-decisions made by the agent as a short temporally ordered list of states, actions and values that can be displayed in a human readable format. The approach relies upon Complementary Learning Systems (CLS) theory to store decisions in episodic memory so that it can be queried to form the explanation. As a result, the approach does not require a second model to be trained post-hoc and can be run on individual episodes. Importantly, the explanations can be used to improve the performance of naive Deep RL agents by simplifying the exploration problem that they face.

\section*{Acknowledgments}

This work was funded by the UK Engineering and Physical Sciences Research Council (EPSRC). We thank NVIDIA for a hardware grant that provided the Graphics Processing Unit (GPU) used to run the simulations.

\urlstyle{rm}
\bibliographystyle{apalike}
\bibliography{EpisodicExplanations}

\newpage
\appendix

\section{Supplementary Materials}

\begin{figure} [H]
	\centering
	\includegraphics[height=17cm, width=14cm]{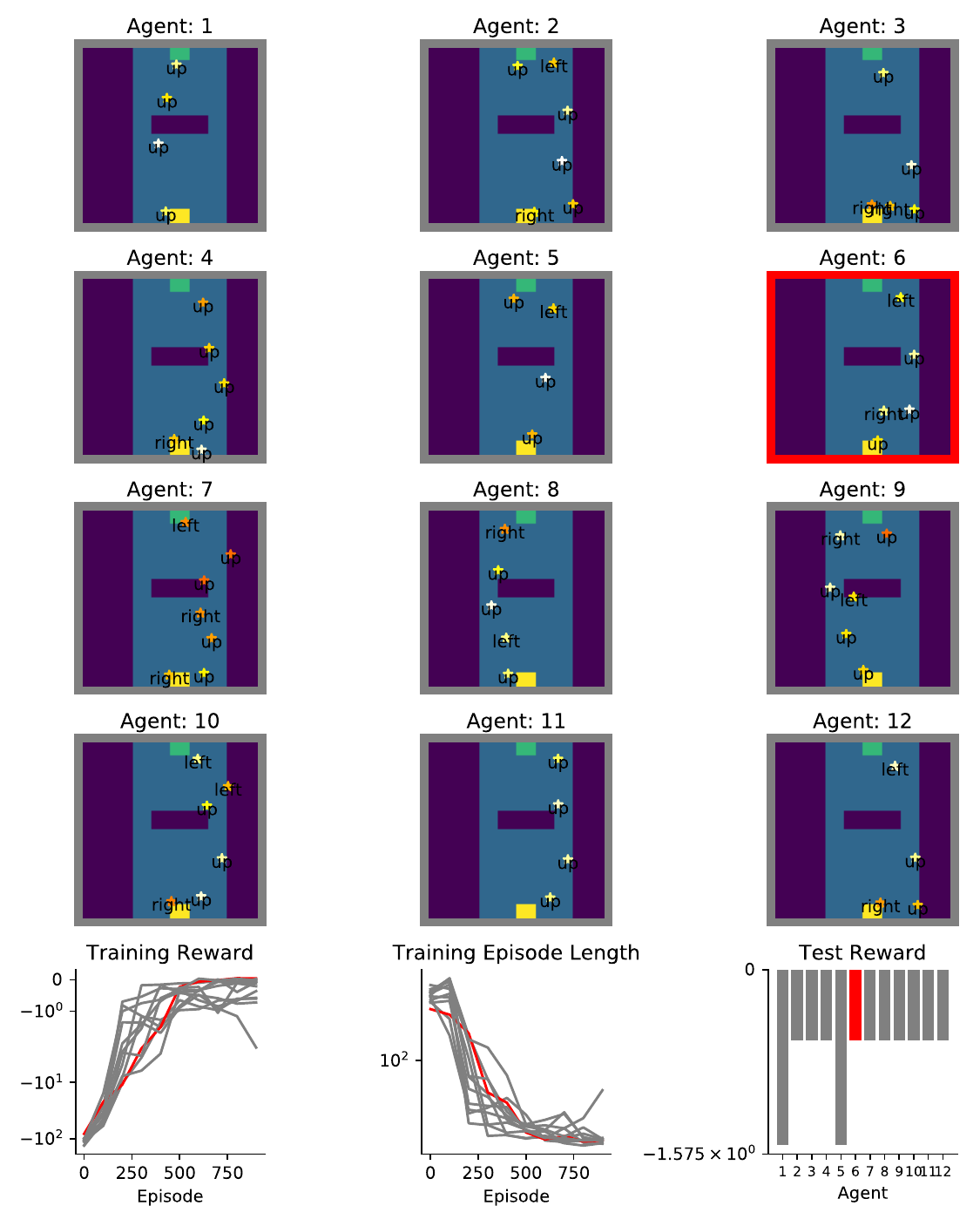}
	\caption{\textit{\textbf{The explanations generated at the end of learning by all 12 agents for the first grid world, including their training and final test performance.} The best performing agent that was used to provide explanations to new agents is highlighted in red. In the case of equal final test reward, the agent with the highest total training reward was chosen as the best performing agent.}}
	\label{fig:grid_world_all_1}
\end{figure}

\begin{figure} [H]
	\centering
	\includegraphics[height=17cm, width=14cm]{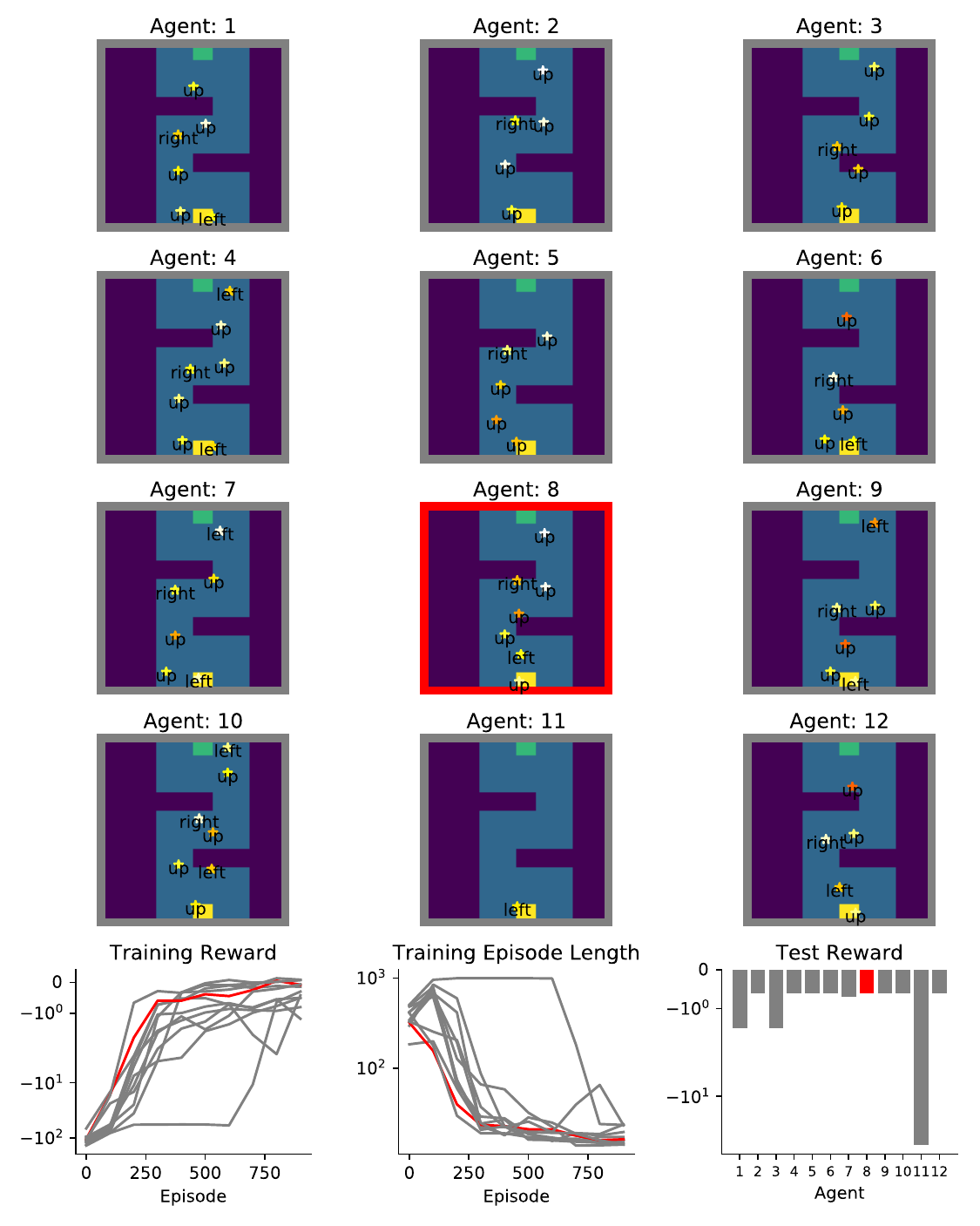}
	\caption{\textit{\textbf{The explanations generated at the end of learning by all 12 agents for the second grid world, including their training and final test performance.} The best performing agent that was used to provide explanations to new agents is highlighted in red. In the case of equal final test reward, the agent with the highest total training reward was chosen as the best performing agent.}}
	\label{fig:grid_world_all_2}
\end{figure}

\begin{figure} [H]
	\centering
	\includegraphics[height=20cm, width=10cm]{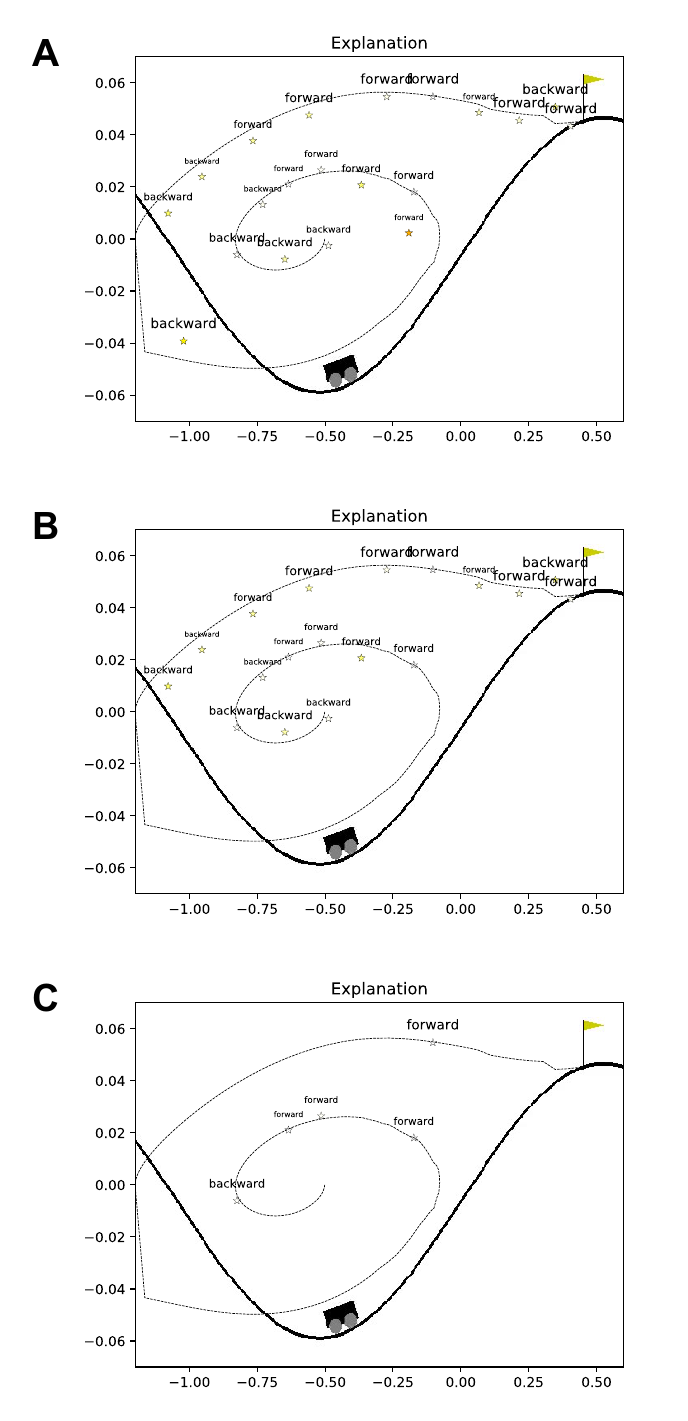}
	\caption{\textit{\textbf{Explanations generated on the continuous mountain car task for different threshold values of $\mathbf{\beta}$ (A: 0.5, B: 0.75, C: 0.99).} Increasing the threshold value reduces the length of the explanation by only incorporating episodic memories that are heavily relied upon during policy execution.}}
	\label{fig:explanation_threshold}
\end{figure}

\begin{figure} [H]
	\centering
	\includegraphics[height=7.25cm, width=14.5cm]{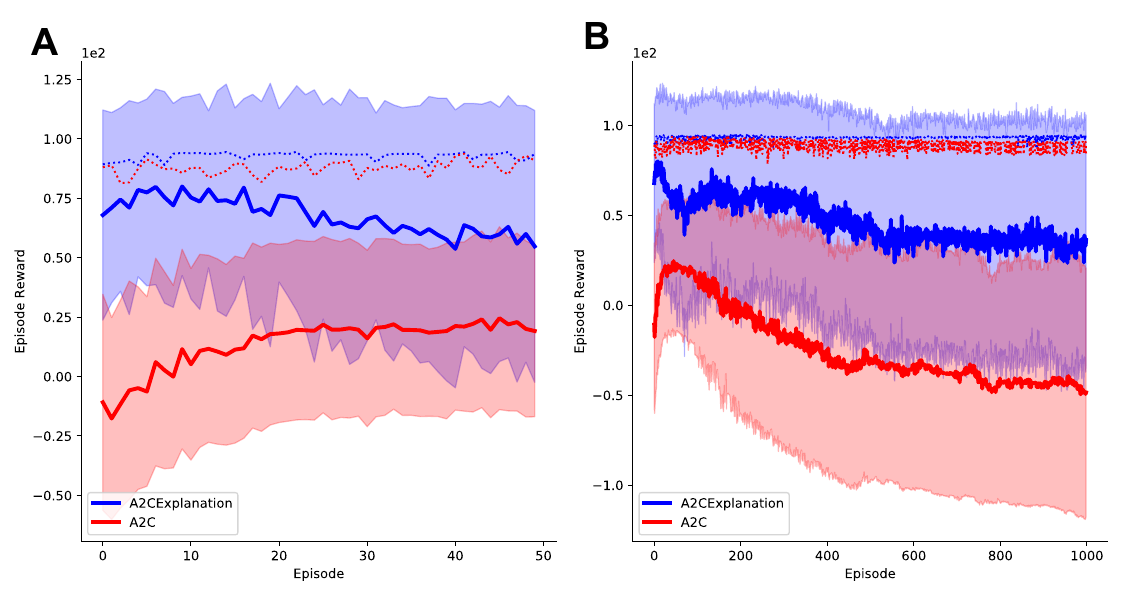}
	\caption{\textit{\textbf{The performance on the continuous mountain car task of A2C agents that did not receive an explanation (A2C) and A2C agents that did receive an explanation generated from CTDL (A2CExplanation).} To generate the explanations, 50 CTDL agents were trained on the continuous mountain car task for 1000 episodes. The agent with the highest total reward on the final episode was then chosen to provide the explanations. The explanations were generated by running the chosen agent on 20 test episodes after training. A2C agents from the Explanation group then picked one of these explanations at random at the start of learning. Solid lines indicate the average performance over 50 agents. Dashed lines indicate the best performing agent for each group. \textbf{(A)} Performance on the first 50 episodes of training. \textbf{(B)} Performance on all 1000 episodes of training.}}
	\label{fig:mountain_car_A2C_comparisons}
\end{figure}

\end{document}